\documentclass[conference]{IEEEtran}
\usepackage{stmaryrd}
\usepackage{amsfonts}
\usepackage{epic}   
\usepackage{ecltree}    
\usepackage{fancybox}   
\usepackage[latin1]{inputenc} 
\usepackage[dvips]{epsfig}

\usepackage{graphicx,times,amsmath,url} 

\IEEEoverridecommandlockouts    

\begin{document}

\title{\ \\ \LARGE\bf SOAP vs REST: Comparing a master-slave GA implementation}
\author{
P.A. Castillo \and J.L. Bernier \and M.G. Arenas \and J.J. Merelo \and P. Garc\'{\i}a-S\'{a}nchez
\thanks{GeNeura. Department of Architecture and Computer Technology. CITIC. University of Granada. {\tt http://geneura.wordpress.com}  Contact email: {\tt pedro@atc.ugr.es} }}

\maketitle

\begin{abstract}

In this paper, a high-level comparison of both SOAP (Simple Object Access Protocol) and REST (Representational State Transfer) is made. These are the two main approaches for interfacing to the web with web services.
Both approaches are different and present some advantages and disadvantages for interfacing to web services: SOAP is conceptually more difficult (has a steeper learning curve) and more ''heavy-weight'' than REST, although it lacks of standards support for security.
In order to test their eficiency (in time), two experiments have been performed using both technologies: 
a client-server model implementation and a master-slave based genetic algorithm (GA).
The results obtained show clear differences in time between SOAP and REST implementations.
Although both techniques are suitable for developing parallel systems, SOAP is heavier than REST, mainly due to the verbosity of SOAP communications (XML increases the time taken to parse the messages).

\end{abstract}

\section{Introduction}
\label{sec:introduction}

Service Oriented Architecture (SOA) \cite{PAPAZOGLOU} is a paradigm for organizing and utilizing distributed computational resources, called services. Using this paradigm, the service providers publish the descriptions (or interfaces) of the services they offer in a service registry, so the service requesters can discover them and bind to the correspondant service provider.
The Web Services are the key point of integration for different applications belonging to different platforms, languages, systems since they are based in a set of standards that make them independent of the underlaying technologies used for providing them.

Although there are several technologies for developing web services (SOAP, REST or XMLRPC among others \cite{compxmlrpc1,compxmlrpc2}), nowadays the main approaches are SOAP (Simple Object Access Protocol) \cite{SOAP,avila01} and REST (Representational State Transfer) \cite{wikiREST}.
Both implementacions are suitable for designing Web Services, however, it is important to understand the pros and cons of each one.

SOAP is the traditional, standards-based approach, but the majority of the web services with public API offer REST interfaces, while some of them offer both REST and SOAP and very few offer just SOAP.
All of the major Web Services providers use REST: Twitter, Yahoo's, Flickr, del.icio.us, pubsub, bloglines, technorati, and several others. Both eBay and Amazon have Web Services for both REST and SOAP.

On the other hand, SOAP Web Services are used in lots of enterprise software as well; for example, Google implements their Web Services using SOAP, with the exception of Blogger, which uses XML-RPC, an early and simpler pre-standard of SOAP. 

The philosophies of SOAP and RESTful Web Services are very different. Strictly, SOAP is a protocol for distributed computing, whereas REST adheres much more closely to a web-based design. 
SOAP requires a greater implementation and understanding effort of the client side to difference of REST based APIs, that focus these efforts on the server side. 
Table \ref{tabla:pro_cons} shows the main strengths and weaknesses for both SOAP and REST.

\begin{table*}[!ht]
\begin{center}
\caption{\scriptsize{Strengths and weaknesses for both SOAP (above) and REST (below).}}
\label{tabla:pro_cons}
\begin{tabular}{|l|l|}

\hline 
\multicolumn{2}{|c|}{SOAP} \\
\hline 
Strengths (pros) & Weaknesses (cons) \\
\hline
+ Handle distributed computing environments & - More verbose \\
+ Built-in error handling  &  - Harder to develop, requires tools \\
+ Extensibility & - Conceptually more difficult, more \\
+ Language, platform, and transport agnostic & \ \ ''heavy-weight'' than REST \\
+ Prevailing standard for web services & \\
+ Support from other standards (WSDL, WS-*) &   \\
\hline

\hline 
\multicolumn{2}{|c|}{REST} \\
\hline 
Strengths (pros) & Weaknesses (cons) \\
\hline
+ Language and platform agnostic & - Assumes a point-to-point communication model \\
+ Much simpler to develop than SOAP & - Not usable for distributed computing environment \\
+ Small learning curve, less reliance on tools & - Lack of standards support for security, etc. \\
+ Concise, no need for additional messaging layer & - Tied to the HTTP transport model \\
+ Closer in design and philosophy to the Web &  \\
\hline

\end{tabular}
\end{center}
\end{table*}

It is important to note that one of the advantages of SOAP is the use of the ''generic'' transport. While REST today uses HTTP/HTTPS, SOAP can use almost any transport to send the request. 
However, one perceived disadvantage is the use of XML because of its verbosity, and the time necessary to parse it.

In this way, in order to determine the efficiency of these two interfacing approaches, we have performed two experiments in which both a SOAP and REST implementations are evaluated:

\begin{itemize}
   \item \textbf{Experiment 1:} a client-server model is implemented, in which the server process runs on a machine and the client processes send and receive text strings.
   \item \textbf{Experiment 2:} a master-slave based GA is implemented, running on the master process the GA and the fitness evaluation on the slave processes.
\end{itemize}

This work continues with our previous research in service oriented
algorithms, as previously stated in \cite{OSGILIATH}, where a
service-oriented platform was presented, or \cite{EVAG}, where studies
about P2P distributed evolutionary algorithms were performed.

This paper is structured as follows:
In sections \ref{sec:SOAP} and \ref{sec:REST} a comprehensive description of SOAP and REST technologies are provided, respectively.
Section \ref{sec:experimento} describes the experiments. In concrete, the client-server and master-slave models implemented for testing are described,
so the experimental configuration, the methodology considered in the study  
; finally, the results obtained are shown.
Last section (Section \ref{sec:conclusionsAndFutureWork}), throws some conclusions and presents the proposed future work.

\section{SOAP: Simple Object Access Protocol}
\label{sec:SOAP}

SOAP is a standard protocol proposed by the W3C (\cite{SOAP}, \cite{avila01}) to interface Web Services, and that extends the remote procedure call (XML-RPC). Thus, SOAP can be considered as an evolution of XML-RPC protocol, much more complete and mature, that allows to perform remote procedure calls to distributed routines (services) based on 
an XML interface as interfacing language.
Thus, SOAP clients can access to objects and metods that are residing in remote servers, using an standard mechanism that makes transparent
the details of implementacion, such us the programming language of the routines, the operating system or the platform used by the provider of the service. At the moment, there exist complete implementations of SOAP for Perl, Java, Python, C++ and other languages \cite{SOAP:soft}.
In opposite to other remote procedure call methods, such as RMI (\emph{remote method invocation}, used by the Java language) or XML-RPC, SOAP has two main advantages: it can be used with any programming language, and it can use any type of transport (HTTP, SHTTP, TCP, SMTP, POP and other protocols).

SOAP sends and receives messages using XML \cite{LearningXML,xml:bible,SOAP:Inside}, wrapped HTTP-in headings.
The interfaces of the metods that can be accessed using SOAP services are specified by a Web Services Description Language (WSDL) \cite{SOAP:WSDL,SOAP:WSDL2}. 
The WSDL of an Web Service consists in an XML description of its interface, i.e., it is a file that describes the name of the methods, the parameters and type of data, the
type of response that the Web Service may return, etc.
Using an WSDL file, that it is based on a neutral language such as XML, the service can be specified for different languages, so that a Java client can access a Perl server.

In this way, SOAP constitutes a high level protocol, making easy the task of distributing objects among different servers, and avoiding the difficulties derived of defining the message formats, nor the explicit call to remote servers.

\section{REST: Representational State Transfer}
\label{sec:REST}

After some years, Internet architects have found an alternative method for building web services in the form of Representational State Transfer (REST)  \cite{wikiREST} .

REST is a style of software architecture for distributed hypermedia systems such as the World Wide Web. The term Representational State Transfer was introduced and defined in 2000 by Roy Fielding in his doctoral dissertation \cite{Fielding2000,Fielding2002}. Fielding is one of the principal authors of the Hypertext Transfer Protocol (HTTP) specification versions 1.0 and 1.1 \cite{wikiREST3,wikiREST4}.

REST-style architectures consist of clients and servers. Clients initiate requests to servers; servers process requests and return appropriate responses. Requests and responses are built around the transfer of representations of resources. A resource can be essentially any coherent and meaningful concept that may be addressed. 

Although REST was initially described in the context of HTTP, is not limited to that protocol. RESTful architectures can be based on other Application Layer protocols if they already provide a rich and uniform vocabulary for applications based on the transfer of meaningful representational state. RESTful applications maximize the use of the pre-existing, well-defined interface and other built-in capabilities provided by the chosen network protocol, and minimize the addition of new application-specific features on top of it.

In a REST environment, clients are not concerned with data storage, which remains internal to each server, so that the portability of client code is improved. 
Servers are not concerned with the user interface or user state, so that servers can be simpler and more scalable. 
Servers and clients may also be replaced and developed independently, as long as the interface is not altered.
Finally, servers are able to temporarily extend or customize the functionality of a client by transferring logic to it that it can execute.

\section{SOAP vs REST: Comparing Efficiency}
\label{sec:experimento}

In this paper we carry out two experiments to compare two parallel models implemented using SOAP and REST technologies in Perl language (due to the familiarity of the authors with this language \cite{perl-ea,optimizing-meta,jjiwann2011}).

The SOAP model was implemented using the {\sf SOAP::Lite}\footnote{http://www.soaplite.com} \cite{SOAP:Lite} module, 
while the REST implementation was carried out using the {\sf Perl Dancer}\footnote{http://perldancer.org} module \cite{perldancer2011,perldancer2011b}, for their stability. In addition, servers developed using these modules are easy to implement and deployed using the computer infrastructure valilable to us in our department.

The Experiment 1 consisted in the implementacion of a client-server model. In this case, the server process runs on a machine that attends client requests, involving different lengths of text strings. 
The experiment 2 implements a master-slave based GA. In this case, a master process runs the GA, while different slave processes evaluate the fitness function.

\subsection{Proof of Concept: Client-Server Efficiency Comparation}

A classic client-server model is implemented in which clients can send and receive a text string.
Different string lengths (100 and 1000 chars) have been configured in order to probe with different loads. In this way, we have tried to determine how the string length (the amount of data) affects the running time (due to communications).

Figures \ref{fig:ejemploSOAP} and \ref{fig:ejemploREST} show the Perl source code of the client-server SOAP and REST implementations.

\begin{figure*}[!ht]
\begin{center}
\setlength{\tabcolsep}{2mm}
\renewcommand{\arraystretch}{1.2}
\begin{tabular}{|cc|cc|}

\hline

\begin{tabular}{l}
 use SOAP::Transport::HTTP; \\
 my \$daemon = \\
 SOAP::Transport::HTTP::Daemon \\ 
 \qquad -$>$ new (LocalPort =$>$ 80)\\
 \qquad -$>$ dispatch\_to('Demo'); \\
 \$daemon-$>$handle; \\
 ~\\
 package Demo; \\
 our \$src=""; \\
 sub push \{ \\
 \qquad my (\$class, \$cad) = @\_; \\
 \qquad \$self-$>$src = \$cad; \\
 \qquad return ''ok''; \\
 \}; \\
 sub pop \{ \\
 \qquad my \$tmp = \$self-$>$src; \\
 \qquad \$self-$>$src = '' ''; \\
 \qquad return \$tmp; \\
 \}; \\
\end{tabular}

& & &

\begin{tabular}{l}
 use Time::HiRes qw( gettimeofday tv\_interval); \\
 use SOAP::Lite; \\ 
 my \$i=0; \\
 my \$tmp\_it = [gettimeofday()]; \\
 for (\$i=0; \$i$<$100 ; \$i++) \{ \\
 \qquad my \$cad=''01234567890 ... 01234567890''; \\
 \qquad print SOAP::Lite \\
 \qquad \qquad -$>$ uri('http://www.soaplite.com/Demo') \\
 \qquad \qquad -$>$ proxy('http://vaio/') \\
 \qquad \qquad -$>$ push(\$cad) -$>$ result; \\
 \qquad print SOAP::Lite \\
 \qquad \qquad -$>$ uri('http://www.soaplite.com/Demo') \\
 \qquad \qquad -$>$ proxy('http://vaio/') \\
 \qquad \qquad -$>$ pop() -$>$ result; \\
 \}; \\
 print ''TIME: '', tv\_interval( \$tmp\_it ); \\
\end{tabular}

\\ 

\hline

\end{tabular}
\caption{SOAP programming example: server (left) and client (right). The string \$cad value varies from 100 to 1000 chars in order to configure different loads.}
\label{fig:ejemploSOAP}
\end{center}
\end{figure*}

\begin{figure*}[!ht]
\begin{center}
\setlength{\tabcolsep}{2mm}
\renewcommand{\arraystretch}{1.2}
\begin{tabular}{|cc|cc|}

\hline

\begin{tabular}{l}
 use Dancer; \\
 my \$src = ""; \\ 
 get '/pop/' =$>$ sub \{ \\
 \qquad my \$tmp = \$src; \\
 \qquad \$src = '' ''; \\
 \qquad return \$tmp; \\
 \}; \\
 get '/push/:cad' =$>$ sub \{ \\
 \qquad my (\$class, \$cad) = @\_; \\
 \qquad \$src = \$cad; \\
 \qquad return ''ok''; \\
 \}; \\
 Dancer-$>$dance; \\
\end{tabular}

& & &

\begin{tabular}{l}
 use Time::HiRes qw( gettimeofday tv\_interval); \\
 use LWP; \\
 my \$nav = new LWP::UserAgent; \\ 
 \$nav-$>$agent("RESTzilla"); \\
 my \$i=0; \\
 my \$tmp\_it = [gettimeofday()]; \\
 for (\$i=0; \$i$<$100 ; \$i++) \{ \\
 \qquad my \$cad=''01234567890 ... 01234567890''; \\
 \qquad my \$rpush = new HTTP::Request GET \\
 \qquad  \qquad =$>$ 'http://127.0.0.1:3000/push/'.''\$cad''; \\
 \qquad my \$upush = \$nav-$>$request(\$rpush); \\
 \qquad my \$rpop = new HTTP::Request GET \\
 \qquad  \qquad =$>$ 'http://127.0.0.1:3000/pop/'; \\
 \qquad my \$upop = \$nav-$>$request(\$rpop); \\
 \}; \\
 print ''TIME: '', tv\_interval( \$tmp\_it ); \\
\end{tabular}

\\ 

\hline

\end{tabular}
\caption{REST programming example: server (left) and client (right). The string \$cad value varies from 100 to 1000 chars in order to configure different loads.}
\label{fig:ejemploREST}
\end{center}
\end{figure*}

The implementacion of this experiment was conducted running the server process on a Ubuntu/Linux machine, while the clients were run on a Windows 7 with the Cygwin\footnote{http://www.cygwin.com} environment.

As string lengths, values of 100 and 1000 characters have been used, in order to test whether the communications time depends on the amount of information sent.
In both cases, the experiment was repeated for 50 times measuring the time spent using ''gettimeofday'' function (in order to achieve a good precision).

As shown in Table \ref{tabla:exp1}, the SOAP version takes more time to complete the communications than the REST implementation.

\begin{table}[!ht]
\begin{center}
\caption{\scriptsize{Results obtained on the first experiment (client-server implementations). SOAP version takes a slightly higher time, although differences between sending a 100 chars string and a 1000 chars string are smaller.}}
\label{tabla:exp1}
\begin{tabular}{|c|c|c|}
\hline 
     &   sending 100 chars  &  sending 1000 chars  \\
\hline
     &                      &                      \\
SOAP &    5.64 $\pm$ 0.17   &    5.83 $\pm$ 0.17   \\
     &                      &                      \\
\hline
     &                      &                      \\
REST &    2.56 $\pm$ 0.10   &    3.45 $\pm$ 0.10   \\
     &                      &                      \\
\hline
\end{tabular}
\end{center}
\end{table}

SOAP version takes a slightly higher time, although differences between sending a 100 chars string and a 1000 chars string are smaller.
REST implementation is faster due to the fact that no extra XML information is sent (that reduces the time taken to parse the messages).

\subsection{Master-Slave based GA Implementation}

In the Experiment 2, we have parallelized a GA following a master-slave model.
We do not intend to innovate in terms of the parallel model, but in the implementation (because implementation matters \cite{jjiwann2011}).

There are many ways to implement a distributed genetic algorithm, one of which is the global paralelization (\emph{farming}), in which, as Fogarty and Huang propose \cite{FogartyHuang}, Abramson and Abela \cite{AbramsonAbela}, or Hauser and M\"anner \cite{HauserManner}, individual evaluation and/or genetic operator application are parallelized. 
A master processor supervises the population and select individuals to mate; then slave processors receive the individuals to evaluate them and to apply genetic operators.

An ideal client-server implementation of a distributed evolutionary algorithm could be a server process with several threads. Each thread would include a population, and would communicate with other threads through the shared code among them. Each thread would use an own tail of individuals to send to other threads. Each thread would evaluate its individuals in different remote computers, carrying out the communication using a REST server.

However, as we cannot use a threaded version of the Perl modules, our implementation will focus on the fitness function evaluation.

\begin{figure}[!ht]
\begin{center}
\epsfig{file=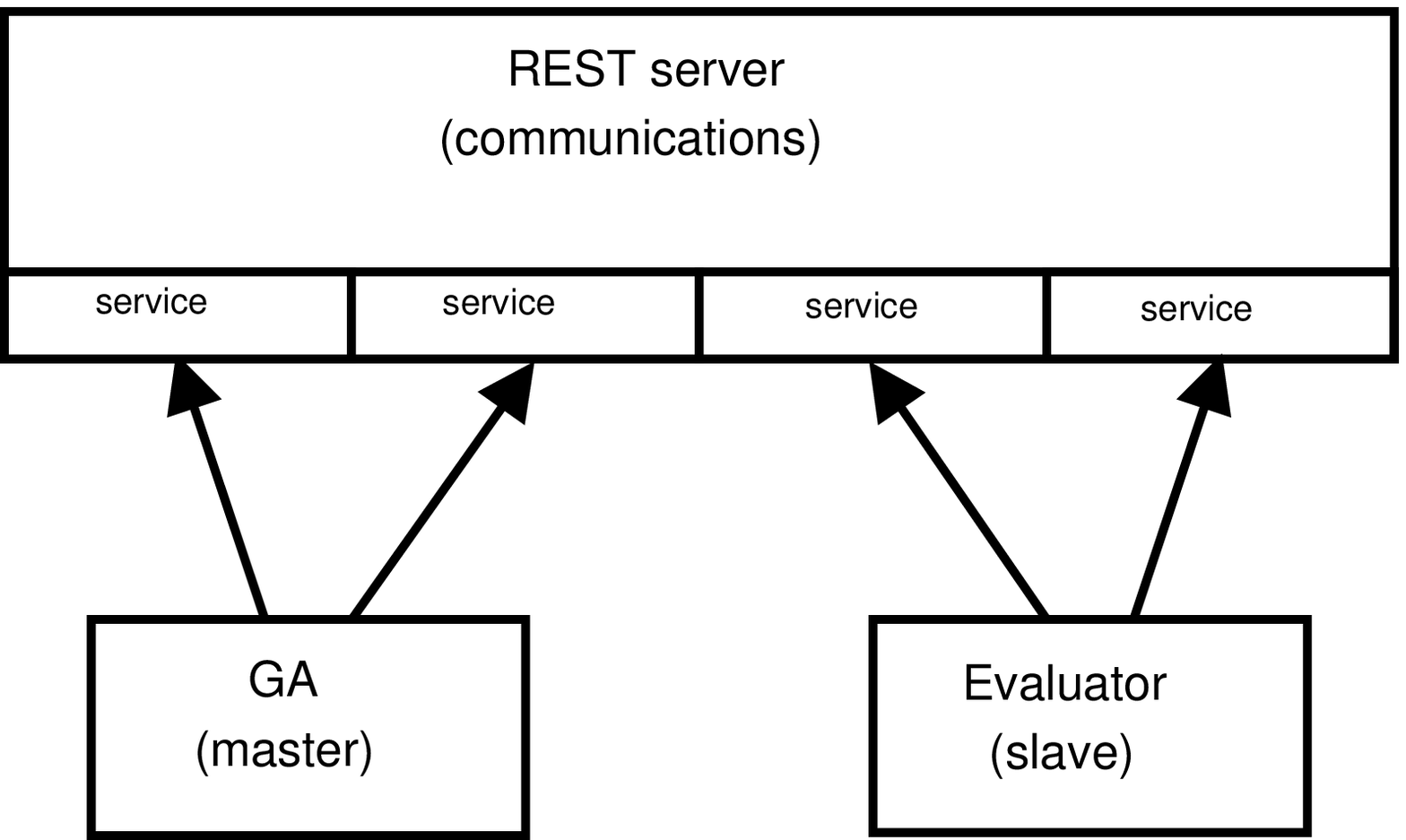,width=9cm}
\caption{Schema of the master-slave based GA implemented in the second experiment. The master process runs the GA and the slave processes evaluate the fitness function.} 
\label{fig:esquema}
\end{center}
\end{figure}

Thus, the simplest way of task distribution along this model is to evaluate the individual fitness function on the clients and to do the other steps on a master process (as shown in Figure \ref{fig:esquema}); this scheme is usually called {\em farming}.

The evolutionary algorithm has been implemented using the Algorithm::Evolutionary (A::E) library \cite{perl-ea,jjSOCO2010}. Version 0.76.2 is used in this work, available at {\sf http://opeal.sourceforge.net } under GPL license.

\begin{figure}[!ht]
\begin{center}
\epsfig{file=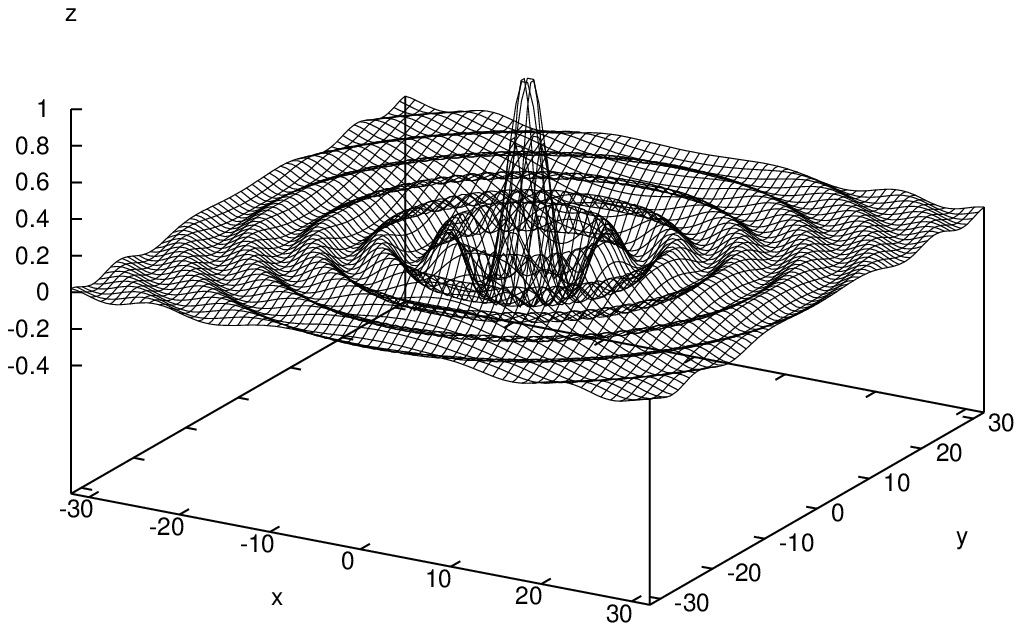,width=7cm}
\caption{Fitness function representation (given by equation \ref{ecu:Marea}). The optimum of this function is $f(0,0)=1$.}
\label{fig:graficaMarea}
\end{center}
\end{figure}

In this experiment, the fitness function is devoted to optimize the function given by equation \ref{ecu:Marea}, which is plotted in Figure \ref{fig:graficaMarea}. Our aim is to find the optimum ($f(0,0)=1$) with an accuracy of $10^{-6}$. 

\begin{eqnarray}
f(x,y)=1+\frac{sin(\sqrt{x^{2}+y^{2}})}{\sqrt{x^{2}+y^{2}}} 
\label{ecu:Marea}
\end{eqnarray}

GA individuals are represented using bitstrings (data type {\sf A::E::Individual::bitstring}). 
As genetic operators, a bitflip mutation ({\sf A::E::Op::Mutation}) and a two points crossover ({\sf A::E::Op::Crossover}) are used.

Remainder GA parameter values are set as follows (default values are used, since we do not intend to find the optimal ones, but to prove feasibility of the implementation, and carry out a comparison):

\begin{itemize}
   \item Population size = 50
   \item Generations = 20
   \item Mutation rate = $20\%$
   \item Crossover rate = $80\%$
   \item Selection rate = $40\%$
\end{itemize}

The full source code (servers, GA and evaluators) and experiment data are available under GPL at: \\  
{\sf http://atc.ugr.es/pedro/RESTvsSOAP.tgz}

As seen in Table \ref{tabla:exp2}, the REST implementation is faster, due to the SOAP verbosity and time taken to decode the XML messages.

\begin{table}[!ht]
\begin{center}
\caption{\scriptsize{Results obtained on the second experiment (master-slave implementations). Both implementations obtain good results using even a small number of generations and population size. As far as the running time is concerned, REST implementation is faster in both configurations (10 gen. / 10 indiv. and 20 gen. / 50 indiv.).}}
\label{tabla:exp2}
\begin{tabular}{|l|r|l|l|}
\hline 
\multicolumn{2}{|c|}{} & 10 generations            &  20 generations   \\
\multicolumn{2}{|c|}{} & 10 individuals            &  50 individuals   \\
\hline
SOAP & accuracy        &  0.997942 $\pm$ 0.000762  & 0.999867 $\pm$ 0.000101 \\
     & time (sec.)     &  3.79 $\pm$ 0.42          & 31.03 $\pm$ 1.89 \\
\hline
REST & accuracy        &  0.996092 $\pm$ 0.004081  & 0.999976 $\pm$ 0.000003 \\
     & time (sec.)     &  2.06 $\pm$ 0.08          & 15.05 $\pm$ 1.17 \\
\hline
\end{tabular}
\end{center}
\end{table}

Both implementations obtain good results in terms of accuracy (both find the optimum with an accuracy of $10^{-6}$) using even a small number of generations and population size. 
As far as the running time is concerned, REST implementation is faster for both load configurations.
As in the client-server experiment, it might be due to the XML verboseness of SOAP communications (that increases the time taken to parse the messages).

\section{Conclusions}
\label{sec:conclusionsAndFutureWork}

As reported in the experiments provided, both techniques are suitable for developing parallel systems.
However, SOAP is heavier than REST, due to the verbosity of SOAP communications (XML increases the time taken to parse the messages).

On another hand, REST technology could not be used to implement a distributed GA following the island model as it does not support asynchronous processing and invocation, while SOAP does support it.

We can conclude that each technology approach has their uses. Moreover, they both have pros and cons. 
However we can devise some applications/situations where one of them might work better than the other:
\begin{itemize}
\item REST is more suitable if...
    \begin{itemize}
		\item bandwidth and resources are limited
		\item stateless \emph{CRUD} (Create, Read, Update, and Delete) operations are needed (operation does not need to be continued)
		\item the information can be cached because of the totally stateless operation of the REST approach
    \end{itemize}	

\item SOAP is a good solution if...
    \begin{itemize}
		\item the application needs asynchronous processing and invocation 
		\item the application needs a guaranteed level of reliability and security
		\item both sides (provider and consumer) have to agree on the exchange format (rigid specifications)
		\item the application needs contextual information and conversational state management (stateful operations)
    \end{itemize}
\end{itemize}

From these REST implementations, several paths for improvement are devised: 
changing the models so that more computation is moved to the clients, leaving the server as just a hub for information interchange among clients; that information interchange will have to be reduced to the minimum. That will make this model closer to the island model, with just the migration policies regulated by the server. That way, the server bottleneck is almost eliminated. 

As future research, it could be of interest adding support for SOAP and REST to existing distributed evolutionary algorithm libraries, such as JEO \cite{maribel:jp2001}, EO \cite{EOEA01}, and libraries in other languages, in order to allow the implementation of multi-language evolutionary algorithms.
In these experiments, Perl language has been used. It could be interesting to test these technologies using other programming languages and libraries (i.e. Java).

\section*{Acknowledgements}

This work has been supported in part by 
the CEI BioTIC GENIL (CEB09-0010) MICINN CEI Program (PYR-2010-13) project, 
the Junta de Andaluc\'{\i}a TIC-3903 and P08-TIC-03928 projects, and 
the Ja\'en University UJA-08-16-30 project.

\end{document}